\documentclass[conference]{IEEEtran}
\IEEEoverridecommandlockouts

\usepackage{cite}
\usepackage{amsmath,amssymb,amsfonts}
\usepackage{algorithmic}
\usepackage{graphicx}
\usepackage{textcomp}
\usepackage{multirow}
\usepackage{algorithm}
\usepackage{algorithmic}
\usepackage{orcidlink}
\usepackage{hyperref}
\usepackage{booktabs}
\usepackage{xcolor}
\def\BibTeX{{\rm B\kern-.05em{\sc i\kern-.025em b}\kern-.08em
    T\kern-.1667em\lower.7ex\hbox{E}\kern-.125emX}}
\begin{document}

\title{Minimum-Sufficient Knowledge and Computation for Wind Energy Predictive Analysis\\
\thanks{Funding Agency: NF25OC0110035 Grant, NNF Denmark}
}

\author{\IEEEauthorblockN{Purbak Sengupta\,\orcidlink{0009-0000-3380-2147}}
\IEEEauthorblockA{\textit{SDU Center for Software Technology} \\
\textit{University of Southern Denmark (SDU)}\\
Vejle, Denmark \\
pur@mmmi.sdu.dk}
\and
\IEEEauthorblockN{Aparna S. Varde\,\orcidlink{0000-0002-3170-2510}}
\IEEEauthorblockA{\textit{SDU Center for Software Technology} \\
\textit{University of Southern Denmark (SDU)}\\
Vejle, Denmark \\
apva@mmmi.sdu.dk}
\and
\IEEEauthorblockN{Emil Jørgensen Njor\,\orcidlink{0000-0003-0219-1624}}
\IEEEauthorblockA{\textit{SDU Center for Software Technology} \\
\textit{University of Southern Denmark (SDU)}\\
Vejle, Denmark \\
njor@mmmi.sdu.dk}
}

\maketitle

\begin{abstract}

Resource-constrained cyber-physical decision systems often apply fixed sensing, knowledge, and computational pipelines despite state-dependent information needs. We introduce Minimum-Sufficient Knowledge and Computation (MSKC), a resource-aware computing principle: select the least-costly route meeting decision-risk and deployment constraints, escalating only when cheaper routes are insufficient. As proof of concept, we isolate adaptive knowledge routing, the decision underlying Knowledge-Guided Machine Learning (KGML), which conventionally injects external knowledge unconditionally rather than only where needed. We instantiate this for wind ramp-event predictive analysis, an edge-oriented decision-support task, evaluating local, relational, and physics-informed routes with a state-aware gate on publicly available real data from two wind farms. We establish conditions for feasibility preservation, cost dominance, and strict expected savings over always-on routing. At a 35.9\% relational-knowledge injection rate, adaptive routing is statistically indistinguishable from always-on injection on the reported metrics while reducing knowledge calls by 64\%, supporting sustainable, Green-IT-aligned deployment. These results support MSKC's premise that uniformly maximal intelligence is unnecessary. This offers intelligent data analysis for big data streams: the analysis itself decides, per state, how much knowledge and computation the data warrant.

\end{abstract}

\begin{IEEEkeywords}
Adaptive knowledge routing, Decision support, Edge AI, Intelligent systems, Green IT,  Resource-aware computing, Smart Cities and Energy, Sustainability
\end{IEEEkeywords}

\section{Introduction}
\label{sec:introduction}
 
Real-time decision support in renewable energy increasingly relies on intelligent, machine-learning-based systems, many of which face tight computational, memory, and communication constraints, particularly at the edge. Because these systems analyze continuous sensor streams from many assets, this is a big data setting where intelligent data analysis must decide not only what the data say but how much knowledge and computation is worth spending to find out. Knowledge-Guided Machine Learning (KGML) improves reliability by incorporating knowledge such as physical relationships, governing equations, or document-derived rules, motivated by black-box models' reliance on large labeled datasets, poor out-of-distribution generalization, and scientifically inconsistent outputs \cite{karpatne2}; in energy management, GIS-derived knowledge improved power plant classification from satellite imagery \cite{austin}. In current practice, however, this knowledge is applied unconditionally, raising computational and communication cost, and with it carbon emissions and sustainability concerns. The Green AI perspective argues this tension should be an explicit evaluation criterion alongside accuracy \cite{schwartz}, and green-computing research has addressed it through data-center decision support balancing efficiency, accuracy, and environmental impact \cite{pawlish_cikm}. Whether the accuracy gain justifies the cost is rarely examined directly.

We introduce Minimum-Sufficient Knowledge and Computation (MSKC), a resource-aware computing principle: a real-time decision system should use the least-costly combination of sensing, knowledge, computation, and execution that meets a prescribed reliability bar and any physical constraints, escalating only when a cheaper option is certified insufficient. For knowledge acquisition, MSKC reframes KGML as adaptive knowledge routing, making knowledge injection a state-dependent decision rather than an always-on property. We test this premise for knowledge acquisition alone: do states exist where extra knowledge is unnecessary, and can they be found without losing decision quality? If so, adaptive sensing, model fidelity, and execution placement, for smart energy systems and beyond, become natural extensions of the same routing problem.

\begin{figure*}[t]
    \centering
    \includegraphics[width=0.9\linewidth]{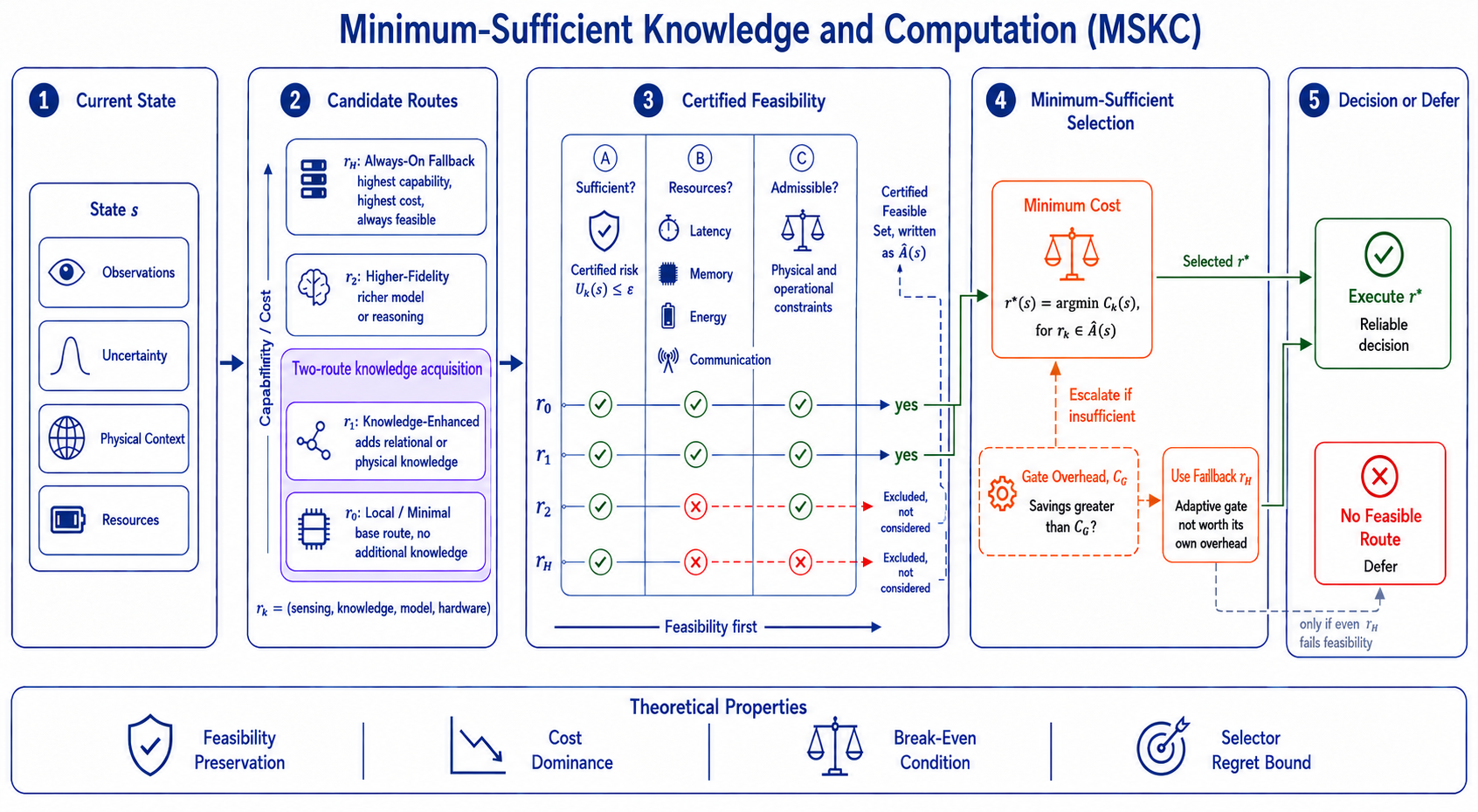}
    \caption{Overview of the MSKC selection process: candidate routes are filtered by certified sufficiency, resource, and admissibility checks into the feasible set $\widehat{\mathcal{A}}(s)$, from which the minimum-carbon route $r^*(s)$ is selected, with fallback to $r_H$ or deferral as described in Section~\ref{sec:mskc}. The highlighted region marks the two-route knowledge-acquisition instance evaluated empirically in this paper.}
    \label{fig:mskc}
\end{figure*}

 
This paper makes four main contributions:

\begin{enumerate}
\item We formulate MSKC, a framework that picks the cheapest knowledge route that is accurate enough and physically valid.
\item We establish its guarantees: it stays within the risk bound when risk estimates are honest, never costs more than the fallback route (strictly less whenever a cheaper sufficient route exists), pays off only when savings exceed routing overhead, and, when the selector is learned and imperfect, loses at most a bounded amount versus the ideal one. In the two-route case, savings grow with how often the richer route is avoided, and once the correct decision region is identified, extra precision adds nothing.
\item We apply part of the framework to wind ramp-event forecasting with two knowledge routes: one reads a neighboring turbine (needs communication), the other uses a local physics signal (needs almost none).
\item We test it on two public wind-farm datasets, asking whether smart routing cuts knowledge use without hurting decisions, beats random or rule-based routing, and lets a lighter model keep up, giving early evidence of minimum sufficiency in both knowledge and computation.
\end{enumerate}

\section{Background and Related Work}
\label{subsec:background}
 
We ground this investigation in wind farm control and forecasting, whose dynamic, spatially coupled, real-time decisions make it well suited to testing MSKC's knowledge-acquisition premise. Coordinated wake steering has established benefits, yet its deployment remains constrained by model fidelity, uncertainty, and the computational cost of running an accurate enough model in real time \cite{wake_steering}. Turbine layout and inter-turbine spatial relationships likewise materially affect energy output and wake behavior, motivating explicit use of a turbine's neighbor states rather than isolated modeling \cite{geospatial_layout}. At the opposite end of the fidelity spectrum, physics-informed surrogate models approximate flow fields under governing laws such as continuity and momentum conservation, trading costly CFD simulation for a physically consistent approximation \cite{physics_surrogate}. Together, these lines of work raise an unresolved question: when should richer relational or physical knowledge be invoked, and when does a turbine's local state already suffice?

Event-based forecasting of wind ramps offers a concrete setting for this question. The original ramp-characterization approach decomposes time-varying wind power into significant and stationary events, extracting features such as duration, magnitude of change, slope angle, and frequency, but was designed for offline event detection and compression rather than online prediction \cite{rbatheta_original}. A later reformulation reversed the usual trajectory-first workflow, predicting an event's semantics before reconstructing its trajectory, as an agentic workflow \cite{rbatheta_agentic}. Our label-generation scheme descends directly from it, except that rather than forecasting an exact trajectory, we predict which of a small number of ordinal severity regions a ramp event falls into.

A separate line of KGML research has pushed toward richer, more integrated knowledge sources for power-grid prediction. Hybrid knowledge-graph and meta-learning approaches have been proposed to improve dynamic prediction of reactive power-voltage response characteristics in renewable-rich power systems \cite{ye_kg}, and large language models to exploit heterogeneous edge data in modern power systems \cite{liang_llm}. Both bring genuine predictive benefit, yet add a cost rarely acknowledged explicitly, precisely the gap MSKC's knowledge-acquisition dimension targets: not whether such sources are useful, since the literature above suggests they generally are, but whether they are necessary for every input. That cost must be weighed against the computational and communication resources a knowledge source consumes once deployed, not judged on predictive accuracy alone. Machine learning systems research argues that data, algorithms, and hardware should be optimized jointly rather than as independent layers, since a statistically accurate model may still be unusable if its resource requirements exceed what the platform can support \cite{ml_systems}. Edge-AI-oriented predictive-maintenance research makes a related argument for industrial edge deployment, showing that reducing a model's input size and sampling rate can simultaneously lower sensing, preprocessing, memory, and inference cost, and that this should be considered from the outset rather than only at final deployment \cite{tinyml_primer}. Green-computing research makes a parallel point for infrastructure: data centers built for peak demand run at low average utilization, wasting energy, and a hybrid design shifting demand spikes to the cloud was found preferable to a purely in-house or purely cloud model \cite{pawlish_sigmod}, later developed into a decision support tool for data center management \cite{pawlish_ijcac}. MSKC adopts this premise, applying it to the choice of knowledge route rather than only to model, input, or infrastructure provisioning.

Finally, MSKC is positioned as complementary to, rather than competing with, other technologies for trustworthy wind-turbine decision support. Bayesian-network-based reliability assessment offers a principled way to represent uncertainty in turbine condition monitoring \cite{bayesian_reliability}, and digital twins are an increasingly common environment for moving from passive monitoring toward adaptive control, though real-time adaptability and interoperability remain open challenges \cite{digital_twin_lifecycle}. Digital-twin-integrated architectures combining temporal convolutional networks with attention have likewise been used to predict turbine runaway from operational SCADA data containing missing values and sensor faults \cite{digital_twin_runaway}. MSKC instead addresses a layer beneath these: given that any of them may draw on additional knowledge or a higher-fidelity model, when is that draw actually worth making?


\begin{figure}
    \centering
    \includegraphics[width=0.6\linewidth]{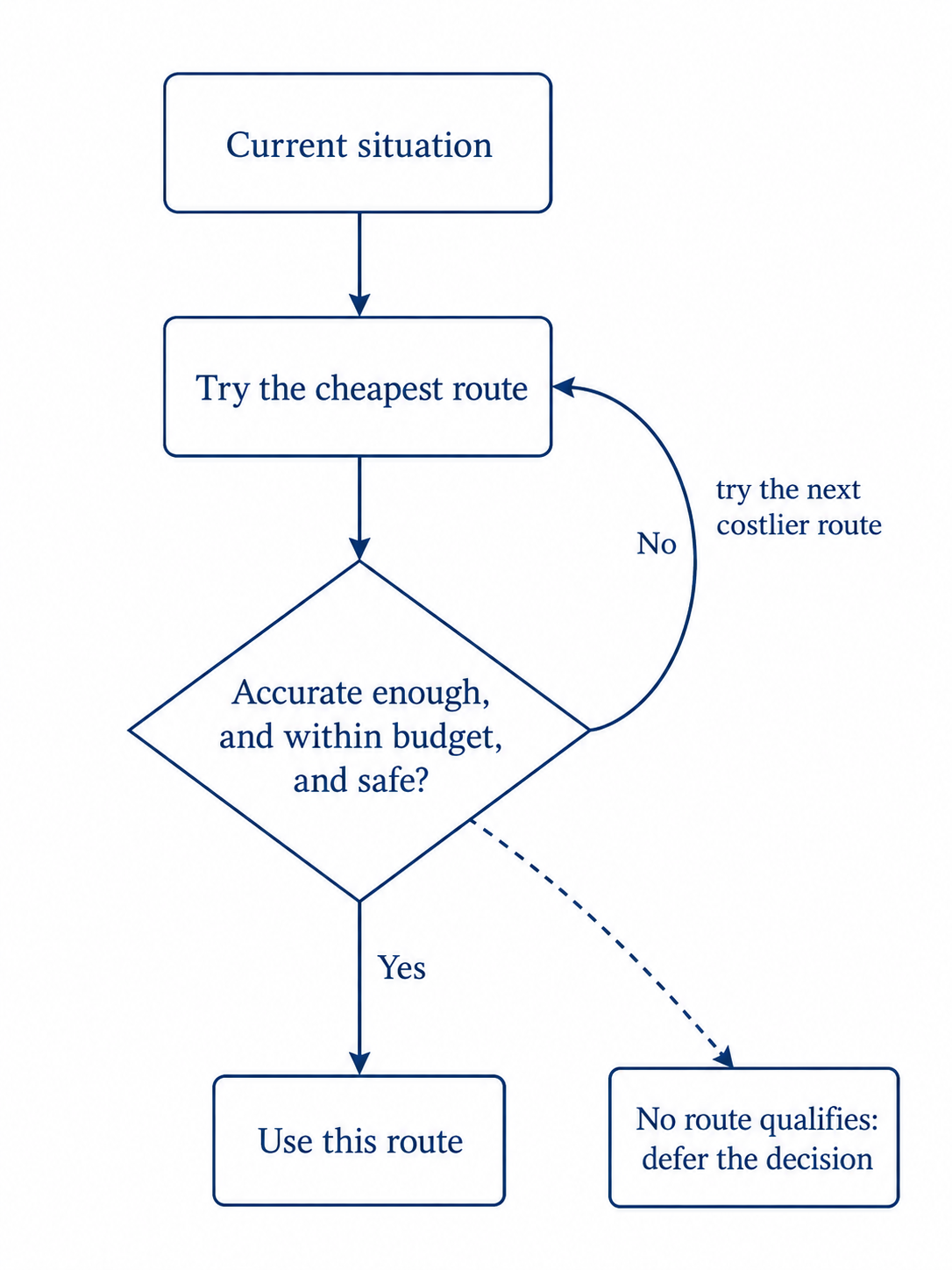}
    \caption{The minimum-sufficient route selection procedure. Starting from the cheapest candidate route, the system checks accuracy, resource fit, and safety together; a failed check escalates to the next costlier route, while a route that passes all three is used immediately. If no route in the menu ever passes, the system defers the decision rather than acting on an unsupported one.}
    \label{fig:decision-flow}
\end{figure}

\section{Proposed Method: Minimum-Sufficient Knowledge and Computation}
\label{sec:mskc}

We answer that question by turning it into a concrete decision procedure: instead of assuming a fixed answer for every situation, the system checks, case by case, whether a cheaper option is already good enough before reaching for something costlier.

\subsection{How the Selector Works}
\label{subsec:plain-terms}

At any moment, a decision system has a menu of different ways it could act, ranging from a cheap, local option to a fully-equipped fallback that is always capable but always costly. We call each of these a \emph{route}: a specific combination of what the system senses, what extra knowledge it consults (if any), which model it runs, and where that model runs, on the device itself, at a nearby edge node, or in the cloud. Figure~\ref{fig:mskc} shows this general picture in full, the complete route menu, all three feasibility checks, and the resulting theoretical guarantees; Figure~\ref{fig:decision-flow} in this section shows the same idea unpacked as the specific, sequential procedure this paper implements. We track all four parts of a route separately, rather than describing a route only by its model, because two routes can share the exact same model while still costing very different amounts: one might require querying a neighboring sensor over a network, another might not, and that difference matters even though the underlying prediction method is identical. MSKC's job is to pick, for the current situation, the cheapest route on that menu that is still good enough to trust.

It does this by asking three questions about each candidate route, in order:

\begin{enumerate}
\item \textbf{Is it likely to be accurate enough?} The true outcome is not yet known, so the system relies on the best available accuracy estimate, which is only useful if it is honest, meaning it never claims a route is safer than it really is.
\item \textbf{Does it fit the resources available right now?} Latency, memory, energy, power draw, and communication are checked separately rather than as one combined score, since a route can be fast but memory-hungry, or cheap to compute but costly to communicate over, and a single score would hide that mismatch.
\item \textbf{Does it obey every hard safety or operational rule?} These are limits on the action itself, such as a regulatory or physical bound, and cannot be traded away for a lower cost.
\end{enumerate}

A route stays in consideration only if it passes all three checks. Passing two is not enough: a route that is accurate and safe but does not fit the available hardware is not a real option. Compactly, the routes that survive all three checks at state $s$ form the \emph{feasible set}
\begin{align}
\widehat{\mathcal{A}}(s)=\{\,r_k:\;& U_k(s)\le\varepsilon,\;\boldsymbol{\rho}_k(s)\preceq\mathbf{b}(s),\nonumber\\
& g_j(a_k(s),s)\le 0\;\;\forall j\,\}.
\label{eq:feasible}
\end{align}
The three conditions match the three questions above, in order. $U_k(s)$ is the estimated risk (expected loss) of route $r_k$ in state $s$, and $\varepsilon$ is the largest risk we are willing to accept. $\boldsymbol{\rho}_k(s)$ lists the route's demands for latency, memory, energy, power, and communication, $\mathbf{b}(s)$ is what the device has available at that moment, and $\preceq$ means every demand fits within its limit. Finally, $a_k(s)$ is the action the route would take, and each $g_j$ is one of the hard safety or operational rules that action must satisfy.

Among the feasible routes, MSKC picks the one with the lowest carbon cost, where $C_k(s)$ denotes the carbon cost of running route $r_k$ in state $s$:
\begin{equation}
r^*(s)=\arg\min_{r_k\in\widehat{\mathcal{A}}(s)} C_k(s).
\label{eq:select}
\end{equation}
Carbon cost is kept distinct from raw energy use, since the same energy can correspond to a different carbon footprint depending on where and when it is spent. If two routes tie, a fixed order over the route menu breaks the tie, so the choice is never left undefined. The menu always includes a designated fallback route $r_H$, assumed feasible in every situation the system is meant to handle, i.e., $r_H\in\widehat{\mathcal{A}}(s)$ for every state $s$; it is not assumed to be the most expensive route in general, only always available as a last resort. If the feasible set is empty, the system does not guess, rather it defers the decision. Figure~\ref{fig:decision-flow} illustrates this decision procedure end to end, including the loop back to a costlier route whenever the current one fails a check, and the deferral path if every route eventually fails.

There is a consequence associated with this design. Because carbon cost is only compared among routes that already passed the accuracy check, a cheap route is never chosen at the expense of being wrong. But the reverse also holds: once a route is judged good enough, MSKC stops looking for something even better, even if a slightly costlier route would have been meaningfully more accurate. MSKC guarantees a quality floor at minimum cost; it does not search for the best possible trade-off point beyond that floor.
Algorithm~\ref{alg:mskc} summarizes this procedure using compact notation: $s$ denotes the current state, $r_0,\ldots,r_K,r_H$ denotes the menu of candidate routes ordered from cheapest to costliest, with $r_H$ the always-available fallback, and $\mathcal{C}$ denotes the set of routes that have passed all three checks (accurate enough, within resource limits, and safe) at that point in the loop.

\begin{algorithm}[t]
\caption{Minimum-Sufficient Route Selection}
\label{alg:mskc}
\begin{algorithmic}[1]
\REQUIRE State $s$; routes $r_0,\ldots,r_K,r_H$ in fixed order; threshold $\varepsilon$; limits $\mathbf{b}(s)$; safety rules $g_j$
\STATE $\widehat{\mathcal{A}} \leftarrow \emptyset$
\FOR{each route $r_k$ in order}
    \IF{$U_k\le\varepsilon$ \AND $\boldsymbol{\rho}_k\preceq\mathbf{b}$ \AND $g_j\le 0\ \forall j$}
        \STATE $\widehat{\mathcal{A}} \leftarrow \widehat{\mathcal{A}} \cup \{r_k\}$ \COMMENT{Eq.~\eqref{eq:feasible}}
    \ENDIF
\ENDFOR
\IF{$\widehat{\mathcal{A}} = \emptyset$}
    \STATE \textbf{defer} the decision
\ELSE
    \STATE $r^* \leftarrow \arg\min_{r_k\in\widehat{\mathcal{A}}} C_k(s)$ \COMMENT{Eq.~\eqref{eq:select}}
    \STATE \textbf{execute} $r^*$
\ENDIF
\end{algorithmic}
\end{algorithm}

\subsection{System Guarantees}
\label{subsec:guarantees-plain}

The construction above gives several properties, each follows directly from how the selection rule is built, not from an additional assumption layered on top.

\textbf{It stays safe, if the accuracy checks are honest:} If the risk estimate $U_k(s)$ in Eq.~\eqref{eq:feasible} never overstates how good a route actually is, then every route that reaches the feasible set genuinely clears the accuracy bar, and whatever the algorithm ultimately selects inherits that same guarantee, since selection only happens among routes that already passed. The system's reliability is therefore never better than the honesty of this estimate; MSKC cannot manufacture safety from an estimate that was already overconfident to begin with.

\begin{figure}
    \centering
    \includegraphics[width=0.9\linewidth]{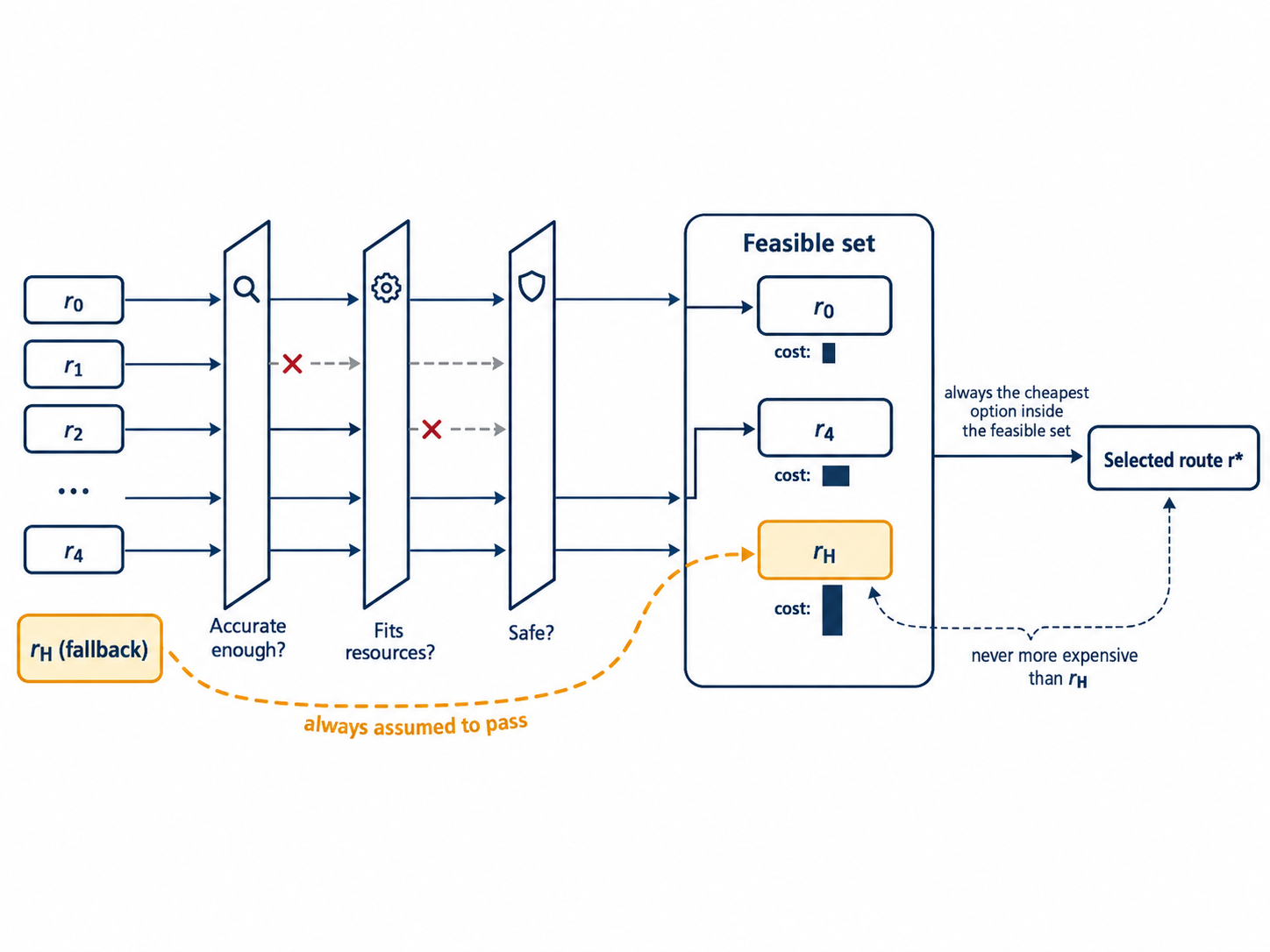}
    \caption{This figure explains why the selected route can never cost more than the fallback $r_H$. Ordinary routes ($r_0, r_1, r_2, \ldots$) must pass all three checks, accurate enough, fits resources, safe, to reach the feasible set; some fail and drop out. $r_H$ is assumed to always pass and reaches the feasible set by a guaranteed path regardless. Because the final selection $r^*$ is always the cheapest option inside this same feasible set, and $r_H$ is always one of its members, $r^*$ can never be more expensive than $r_H$.}
    \label{fig:dominance}
\end{figure}

\textbf{It never costs more than always using the safe fallback:} Because $r_H\in\widehat{\mathcal{A}}(s)$, the fallback is always one of the candidates that Eq.~\eqref{eq:select} chooses among, and since the selected route has the lowest cost in that set,
\begin{equation}
C_{r^*(s)}(s)\;\le\;C_H(s)\qquad\text{for every state } s.
\label{eq:dominance}
\end{equation}
Here $C_{r^*(s)}(s)$ is the carbon cost, in state $s$, of the route $r^*(s)$ that Eq.~\eqref{eq:select} selects, and $C_H(s)$ is the carbon cost in that same state of using the fallback $r_H$ instead, so the result can never be pricier than simply always using the fallback. Figure~\ref{fig:dominance} makes this concrete: the fallback route always reaches the feasible set by a guaranteed path, regardless of whether other routes pass or fail the three checks, so it is always available as a comparison point, and the final selected route, drawn from the same set, is never more expensive than it.

\textbf{It usually costs less:} Whenever a cheaper route is genuinely good enough for at least some real situations, the algorithm will find and use it in those situations rather than defaulting to the fallback out of caution, producing a real, not just theoretical, saving on average across many decisions.

\textbf{The saving has to outweigh the cost of checking:} Running the three checks above is not free: evaluating an accuracy estimate, measuring resource fit, and confirming safety rules all consume some resources of their own. Adaptive routing is only worthwhile once the savings it produces exceed this checking cost; a correct but overly expensive checking procedure can still be a net loss compared to simply always running the fallback and skipping the checks altogether.

\textbf{An imperfect, learned version of this procedure has bounded extra cost, no automatic safety guarantee:} A real system typically approximates the accuracy check with a trained estimator rather than a perfectly calibrated one, since building a fully certified estimate for every situation in advance is often impractical. The resulting extra cost, from occasionally picking a different route than the ideal procedure would have, is bounded by how often the approximation disagrees with the ideal choice, so an approximation that rarely disagrees can only waste a small, bounded amount. This bound is about cost efficiency only; it does not by itself guarantee the approximate version stays safe, since a cost-efficient approximation can still occasionally be overconfident about accuracy. Safety, in that case, has to be checked separately from efficiency, not assumed to follow from it.

MSKC can also justify using a coarser prediction, not just a cheaper route, whenever only the general category of an outcome matters for the action taken, not its exact numeric value. Once the correct category is identified, extra numerical precision cannot change the action, and therefore adds no decision-relevant value. This is a different, weaker kind of sufficiency than the classical statistical notion, which asks whether a summary of the data preserves everything needed to estimate an unknown quantity \cite{casella_berger}; here we only ask whether it preserves the action taken, not the full statistical picture. This paper does not evaluate this mechanism; it is noted here as a complementary direction.

\subsection{Applications and Trade-offs}
\label{subsec:applications}

The routing procedure above is general enough to apply beyond wind ramp forecasting. We outline three illustrative scenarios, each trading off differently along the same accuracy-versus-cost dimension explored empirically in this paper.

The first example is of a wearable device that might run a lightweight, on-device heart-rhythm classifier continuously, escalating to a more detailed cloud-based analysis only when the local classifier's confidence is low. The trade-off here is battery life and user privacy against the risk of a missed detection; unlike the wind-turbine case, an under-escalation here can have direct safety consequences for a person, not just financial cost, so the accuracy threshold would need to be set considerably more conservatively.

Next scenario can be a mobile robot or a vehicle, which might rely on camera-only perception by default, fusing in LIDAR or radar data only when camera-based confidence drops below a safe margin. The trade-off is latency and onboard compute against the same kind of safety-critical miss discussed in this paper's own Discussion section, making this a domain where the false-safe rate, not overall accuracy, would again be the metric to optimize for.

Another example is a factory sensor that flags a component as healthy from vibration data alone, pulling in additional acoustic or thermal sensor streams only when the vibration signal is borderline. Here the trade-off is closer to the wind-turbine case; i.e, the main cost of over-escalating is wasted inspection effort, while under-escalating risks an undetected failure, making it a reasonable candidate for directly reusing the framework and evaluation approach demonstrated here.

Across all three, the accuracy threshold and the specific cost being minimized (energy, latency, communication, or a mix) would need to be chosen based on how severe an under-escalation is in that particular domain. The same trade-off applies to where knowledge itself resides.~\cite{edgekb} envision compact edge knowledge bases (EdgeKBs), built by slicing existing knowledge bases, tailoring new ones, or adapting them at runtime, that keep frequently needed knowledge on-device and retrieve the rest from fog or cloud repositories when required. In MSKC terms, such an EdgeKB is a cheap local knowledge route and the remote repository a costlier one, so a gate of the kind evaluated here could decide when local knowledge suffices.


\section{Intelligent Data Analysis using MSKC}
\label{sec:data-analysis}

\subsection{Data Description in Wind Energy}
\label{subsec:data-description}

We evaluate MSKC on ten-minute Supervisory Control and Data Acquisition (SCADA) data from two real, publicly available onshore wind farms, Kelmarsh \cite{kelmarsh} and Penmanshiel \cite{penmanshiel}, covering the 2024 calendar year. All six Kelmarsh turbines are included; for Penmanshiel, five of its fourteen turbines had usable telemetry in this release. Table~\ref{tab:data-summary} lists farm specifications and post-cleaning data volume. Two data issues were fixed before labeling. First, the raw files contained a duplication bug, about forty repeated rows per real timestamp, resolved by keeping the row with a non-null power reading (zero conflicts across all eleven turbines). Second, near-zero power at high wind speed was turbine downtime rather than curtailment in 96\% of cases, so downtime periods are excluded when building ramp labels. A clear sensor fault was also removed: a single turbine's wind speed jumping to 60~m/s within forty minutes with no matching reading at any neighboring turbine. All other high-wind readings were corroborated by at least one nearby turbine and kept.

\begin{table}[t]
\centering
\caption{Wind farm specifications and dataset summary after cleaning. Ranges are taken across each site's turbines.}
\label{tab:data-summary}
\footnotesize
\setlength{\tabcolsep}{4pt}
\begin{tabular}{lcc}
\toprule
 & Kelmarsh & Penmanshiel \\
\midrule
Turbines & 6 & 5 (of 14) \\
Rated power (kW) & 2050 & 2050 \\
Rotor diameter (m) & 92.0 & 82.0 \\
Hub height (m) & 68.5--78.5 & 59.0 \\
Period & \multicolumn{2}{c}{2024 (1 year)} \\
Resolution & \multicolumn{2}{c}{10 min} \\
Raw rows / turbine & $\approx$2.17M & $\approx$2.45M \\
Clean rows / turbine & 51{,}868--52{,}604 & 52{,}043--52{,}597 \\
Missing timestamps & 0.19--1.6\% & 0.20--1.25\% \\
Cross-turbine overlap & 97.9\% & 98.0\% \\
\bottomrule
\end{tabular}
\end{table}

Ramp labels use five ordinal severity classes, based on the change in normalized power thirty minutes ahead, assigned only over clean, gap-free, downtime-free stretches. Table~\ref{tab:label-distribution} shows the resulting class balance; 95.6\% of the full year, across all turbines, receives a valid label.

\begin{table}[t]
\centering
\caption{Ordinal ramp-region label distribution, pooled across both sites.}
\label{tab:label-distribution}
\footnotesize
\begin{tabular}{lc}
\toprule
Region & Share of labeled rows \\
\midrule
Severe down-ramp & 1.7\% \\
Moderate down-ramp & 10.7\% \\
Stable & 75.1\% \\
Moderate up-ramp & 10.8\% \\
Severe up-ramp & 1.7\% \\
\bottomrule
\end{tabular}
\end{table}

\begin{figure*}[t]
\centering
\includegraphics[width=0.75\textwidth]{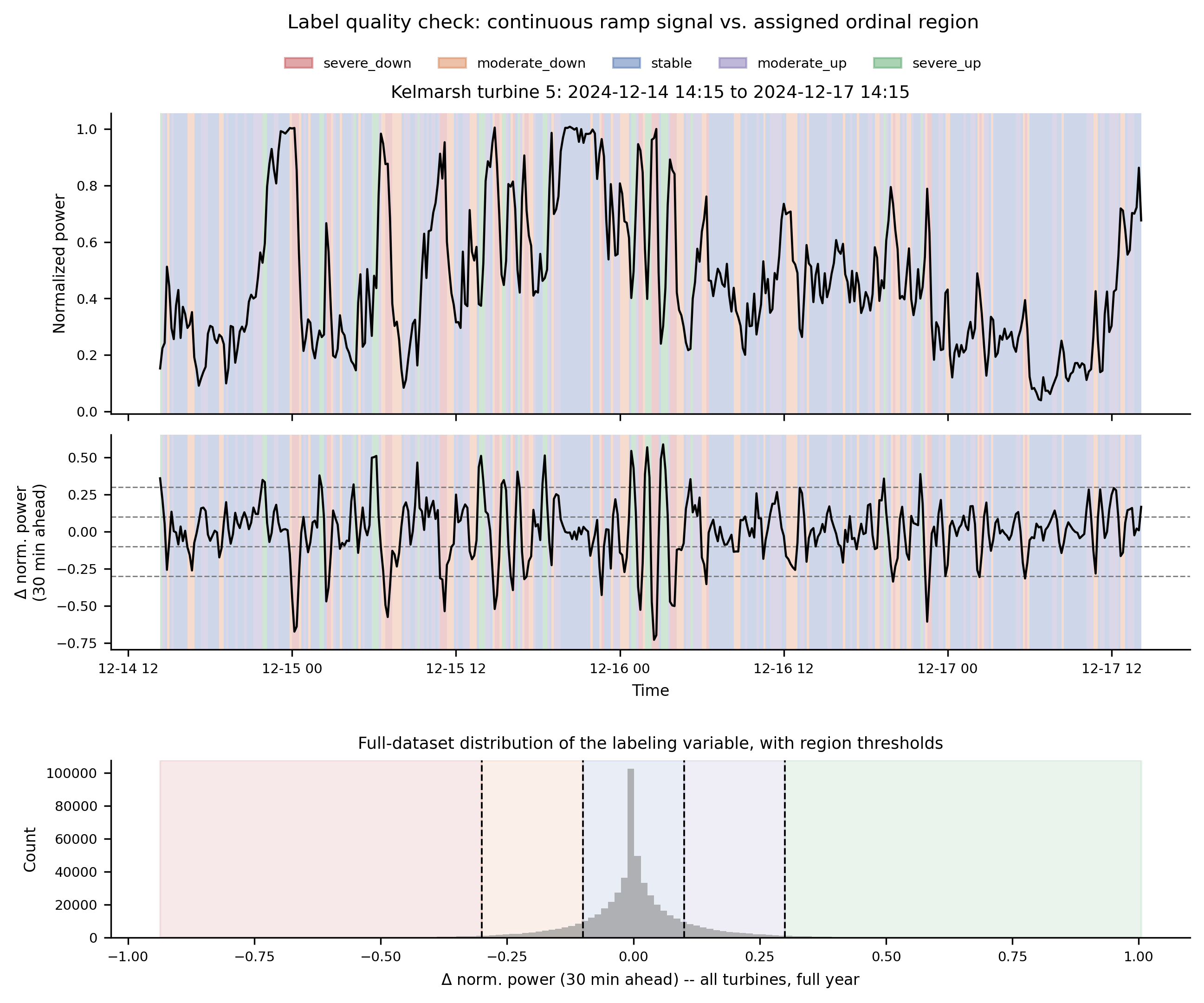}
\caption{Validation of the ordinal ramp-region labeling mechanism: power trace shaded by assigned region (top), the underlying change signal against its thresholds (middle), and the full-year, all-turbine distribution of that signal (bottom).}
\label{fig:label-quality}
\end{figure*}

We also checked the labeling scheme directly against real data, since class balance alone cannot rule out a systematic offset between the signal and its assigned label. Figure~\ref{fig:label-quality} shows a seven-day window, automatically selected as the span with the best mix of severe up- and down-ramp events across all turbines, zoomed to the closest such pair. It plots the power trace shaded by assigned region, the underlying change signal against its thresholds (the shading changes exactly where the signal crosses one), and the full-year distribution of that signal, whose thresholds fall in reasonable, non-degenerate parts of it. Because a label looks thirty minutes ahead, shading can begin slightly before a move finishes in the power trace; this is expected, not an error.

\begin{figure}[t]
\centering
\includegraphics[width=0.75\linewidth]{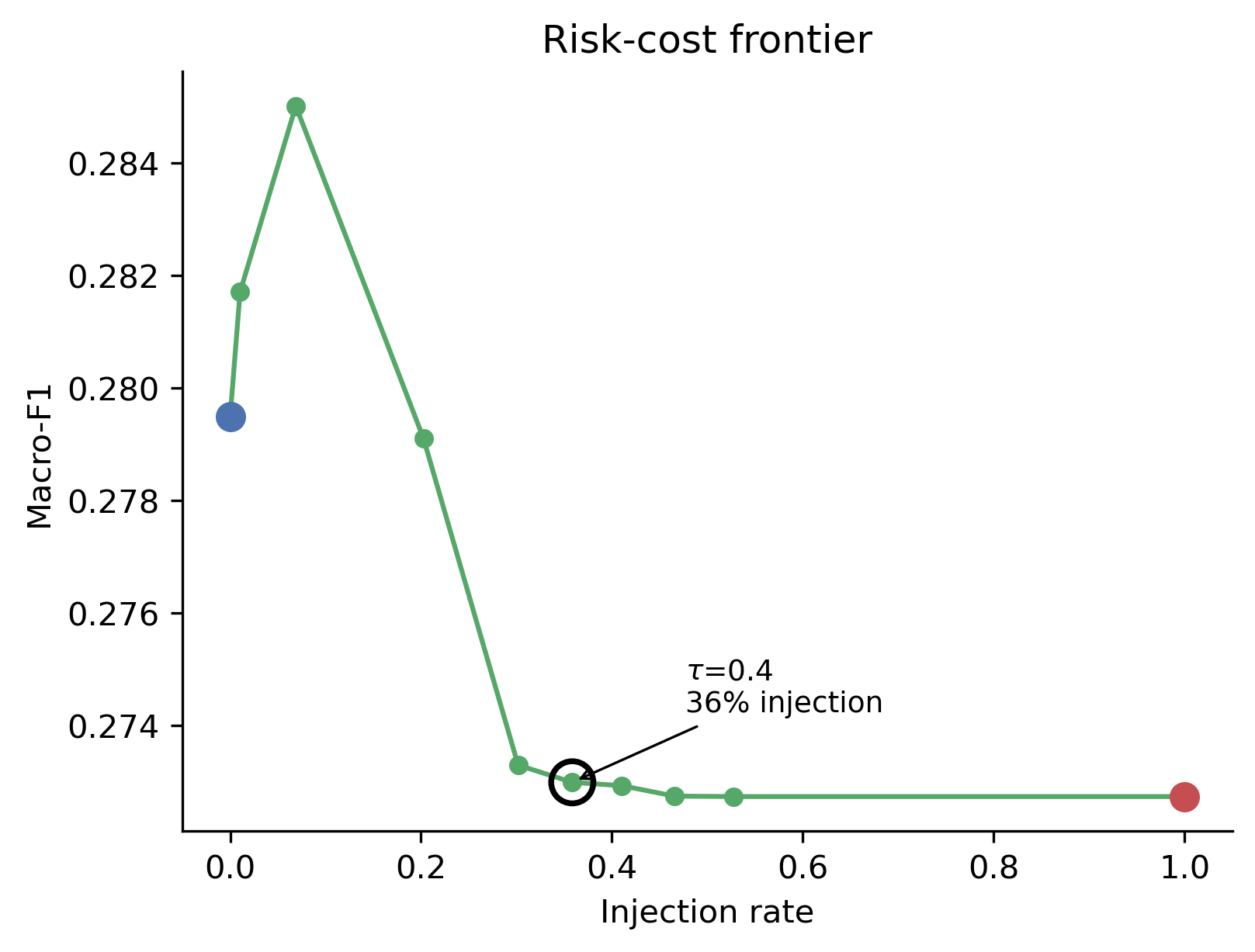}
\caption{Macro-F1 as a function of injection rate under the swept gate threshold.}
\label{fig:risk}
\end{figure}

\begin{figure*}[t]
\centering
\includegraphics[width=0.92\linewidth]{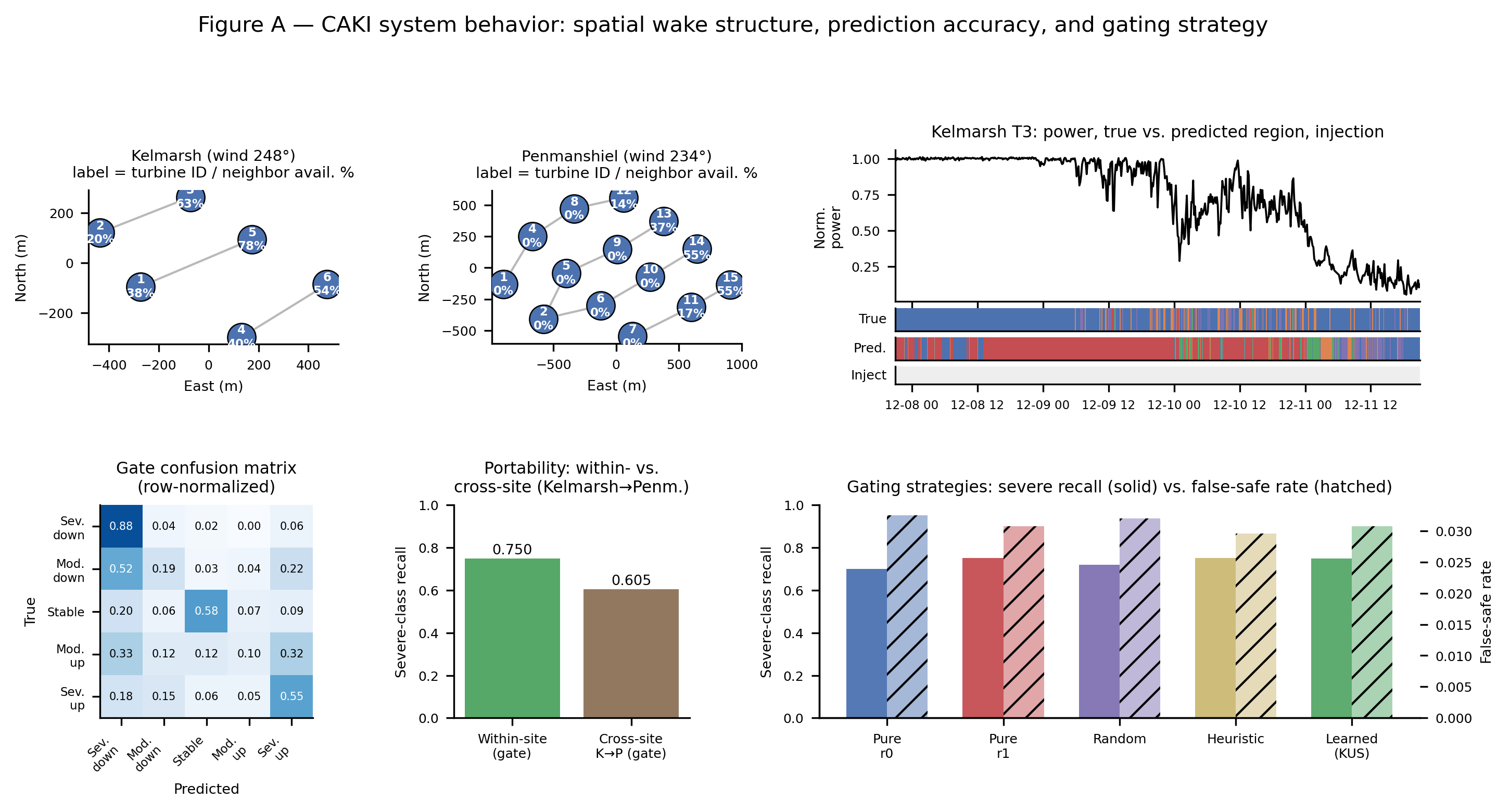}
\caption{Spatial layout and neighbor availability at both sites; a four-day trace of true region, predicted region, and injection decision; the deployed gate's confusion matrix; within-site versus cross-site severe recall; and the learned KUS gating-strategy comparison at matched injection rate.}
\label{fig:system}
\end{figure*}

\subsection{Implementation of MSKC}
\label{subsec:implementation}

A route combines four things: what the system senses, what extra knowledge it consults, which model it runs, and where it runs. The full MSKC vision lets all four vary jointly; this paper varies only the knowledge source, holding the rest fixed so that any difference in decision quality or cost is attributable to knowledge alone. This is the knowledge-acquisition specialization of Algorithm~\ref{alg:mskc}, the narrowest nontrivial slice of MSKC that can still be tested empirically. Because the richer route is assumed to cost strictly more than the base route at every state, each state where the gate declines it avoids that extra cost, so the fraction of declined states is a direct measure of the saving achieved.

We implement three routes. The base route uses only a turbine's own recent SCADA history. The relational route adds the simultaneous reading of the nearest upwind neighbor, determined from turbine coordinates and measured wind direction, which requires cross-turbine communication. The physics-informed route adds the residual between observed power and a power curve fit to that turbine on training data only, a local lookup needing no communication. All three share the same classifier family and execution tier, so only the knowledge source and input features differ. Only the base and relational routes enter the gated decision; the physics-informed route is compared directly against the base route, since its negligible cost leaves little for a gate to weigh.

No literal fallback route of higher fidelity is instantiated. Cost dominance (Eq.~\eqref{eq:dominance}) is stated relative to a fallback, and in the gated comparison always using the relational route plays that role: when Section~\ref{sec:data-analysis} reports the gate matching ``always-on injection,'' this is the baseline meant. A minimal secondary probe of the model dimension is also reported: the base and relational routes are retrained at three classifier weights, holding knowledge fixed, to test whether a lighter model trades away quality independently of the knowledge question.

The implemented selector approximates the ideal rule of Eq.~\eqref{eq:select}, in the sense of the regret guarantee of subsection~\ref{subsec:guarantees-plain}: rather than certifying the risk bound $U_k(s)\le\varepsilon$ of Eq.~\eqref{eq:feasible}, a classifier trained on out-of-bag predictions from both routes estimates, for a given state, whether the knowledge-enhanced route would correct the base route's prediction, with no additional labeled data. Because this score is learned rather than certified, the safety guarantee is not claimed to hold; it would for a selector with a genuinely calibrated bound, for instance one built via conformal prediction. Whether the gate finds sufficient states well enough without that guarantee is the empirical question of subsection~\ref{subsec:experiments}. Its decision threshold is swept from always-injecting to never-injecting, tracing the full tradeoff between injection frequency and decision quality. The reported operating point was selected on a validation slice (the six weeks preceding the test period) as the lowest injection rate whose false-safe rate stays within about one missed severe event of always-on injection, and was then evaluated once on the held-out test set. Code, preprocessing scripts, and a dataset manifest are available at \href{https://github.com/PurbakSengupta/MSKC}{this GitHub link}.

\subsection{Experiments with Results}
\label{subsec:experiments}

All routes are trained on a chronological split, data before 1 November 2024 for training, the remainder held out for testing. Because an upwind neighbor is not available at every state, $r_1$ can only be evaluated on the subset of test rows where one exists (``Neighbor-avail.''); $r_0$ is reported against both this same restricted subset, for a fair paired comparison with $r_1$, and separately against the full test set, for a fair paired comparison with $r_2$, which requires no neighbor and is evaluated on every test row (``Full test''). Route scores are reported in Table~\ref{tab:routes}.

\begin{table}[t]
\centering
\caption{Knowledge route scores, reported separately for each route's own paired evaluation population.}
\label{tab:routes}
\footnotesize
\setlength{\tabcolsep}{3.5pt}
\begin{tabular}{llccc}
\toprule
Population & Route & Macro F1 & Sev. recall & Comm. cost \\
\midrule
\multirow{2}{*}{Neighbor-avail.} & Base ($r_0$) & 0.280 & 0.701 & none \\
 & Relational ($r_1$) & 0.273 & 0.751 & required \\
\midrule
\multirow{2}{*}{Full test} & Base ($r_0$) & 0.263 & 0.694 & none \\
 & Physics ($r_2$) & 0.266 & 0.705 & none \\
\bottomrule
\end{tabular}
\end{table}

Table~\ref{tab:bootstrap} reports paired bootstrap comparisons ($N=1000$ resamples) for each route against its corresponding base row using three metrics: macro F1 (the unweighted mean of the F1 scores across all five ordinal regions), severe recall (the fraction of true severe-class events, upward or downward, correctly identified as severe), and false-safe rate (the fraction of true severe-class events incorrectly predicted as stable, representing the operationally relevant miss). Sweeping the gate decision threshold from always injecting to never injecting the relational route yields the frontier shown in Figure~\ref{fig:risk}. At an injection rate of $35.9\%$, a paired bootstrap comparison with the always-on injection setting ($N=1000$ resamples) gives a macro-F1 difference of $+0.0002$ [$0.000$, $+0.001$], a severe-recall difference of $-0.0013$ [$-0.005$, $+0.002$], and a false-safe-rate difference of $+0.0000$ [$0.000$, $0.000$].

\begin{table}[t]
\centering
\caption{Paired bootstrap comparisons, 95\% confidence intervals.}
\label{tab:bootstrap}
\footnotesize
\begin{tabular}{lccc}
\toprule
Comparison & Macro F1 & Sev. recall & False-safe \\
\midrule
$r_1 - r_0$ & $-0.0068$ & $+0.0511$ & $-0.0017$ \\
 & [$-0.011$,$-0.003$] & [$+0.034$,$+0.068$] & [$-0.008$,$+0.005$] \\
$r_2 - r_0$ & $+0.0036$ & $+0.0115$ & $-0.0009$ \\
 & [$+0.002$,$+0.006$] & [$+0.003$,$+0.020$] & [$-0.004$,$+0.003$] \\
\bottomrule
\end{tabular}
\end{table}

Table~\ref{tab:baselines} compares the learned gate against uninformed random gating (200 trials) and a hand-designed heuristic, at the matched 35.9\% injection rate. Reducing the classifier from 300 to 15 trees changes measured single-row inference latency from 12.0~ms to under 2~ms. Table~\ref{tab:modelweight} reports macro-F1 and severe recall at each weight. Figure~\ref{fig:system} reports spatial layout and neighbor availability at both sites, a four-day trace of true region, predicted region, and injection decision for one turbine, the gate's confusion matrix, a within-site versus cross-site comparison (Kelmarsh trained, Penmanshiel tested), and the gating-strategy comparison from Table~\ref{tab:baselines}.

\begin{table}[t]
\centering
\caption{Gating strategies at matched injection rate.}
\label{tab:baselines}
\footnotesize
\begin{tabular}{lccc}
\toprule
Strategy & Macro F1 & Sev. recall & False-safe \\
\midrule
Random (mean) & 0.277 & 0.719 & 0.032 \\
Heuristic & 0.273 & 0.752 & 0.030 \\
Learned gate & 0.273 & 0.750 & 0.031 \\
\bottomrule
\end{tabular}
\end{table}

\begin{table}[t]
\centering
\caption{Model weight versus latency and decision quality (base route).}
\label{tab:modelweight}
\footnotesize
\begin{tabular}{lccc}
\toprule
Trees & Latency (ms) & Macro F1 & Sev. recall \\
\midrule
300 & 12.0 & 0.327 & 0.53 \\
50 & 2.2 & 0.299 & 0.65 \\
15 & $<$2 & 0.280 & 0.70 \\
\bottomrule
\end{tabular}
\end{table}

Cross-site severe recall is 0.750 within-site and 0.605 when trained on Kelmarsh and tested on Penmanshiel without re-tuning the gate threshold. A compact neural network trained on the same feature sets gives, on the matched neighbor-available population, macro-F1 0.310 ($r_0$) and 0.304 ($r_1$), severe recall 0.548 ($r_0$) and 0.634 ($r_1$).

\section{Evaluation and Discussion}
\label{sec:discussion}

The relational route in Subsection~\ref{subsec:experiments} trades a small decrease in macro-F1 for a larger increase in severe-class recall, while the physics-informed route improves both. We argue recall, not macro-F1, is the metric that matters for the relational trade-off specifically, since the two error types it involves are not symmetric in consequence. A missed severe event triggers no corrective action, no reduced-load posture, no operator alert, and risks unplanned mechanical stress up to structural failure; a false alarm costs a period of curtailed output and a bounded, reversible revenue loss. One view holds that a missed event is categorically worse, since it is the only error type with a plausible path to damage, and that the system should be biased toward recall even at a real cost to precision. A second view does not dispute this ordering but questions whether a correctly-recalled prediction is sufficient in practice: if severity is forecast within a thirty-minute horizon but the event materializes several hours later, an operator may take precaution, observe nothing within a plausible waiting window, stand down, and be unprepared when the delayed event arrives, a true positive by the recall metric but an operational near-miss in reality. We do not resolve this empirically, since it requires modeling operator response and event-timing jointly, which this proof of concept does not attempt, and report it as an open limitation of evaluating a single-horizon classifier against a continuous, imprecisely-timed process. Given how rare severe events are (1.7\% of labeled rows in each direction), we adopted a deliberately modest bar for this proof of concept, that reliably identifying roughly half of severe events without an unacceptable false-safe rate already has real operational value, and the measured recall of 0.70 to 0.75 across all reported routes and gating strategies exceeds it, though we state this as a domain-appropriate target rather than a claim of strong absolute performance.

This connects to a broader pattern in KGML practice: knowledge sources, whether a knowledge graph, a physics-informed constraint, or a neighboring sensor's reading, are typically consulted unconditionally once a system is designed to use them, with that decision made once at design time rather than per input. Our results do not argue against KGML's premise that external knowledge improves decisions, Table~\ref{tab:bootstrap} confirms that it does for both routes evaluated here, but they do show that applying it uniformly is a design choice, not a requirement: the gate identifies, on a per-input basis, a substantial share of states for which the additional knowledge is not needed, at no statistically detectable cost to the decision quality that knowledge was introduced to improve. MSKC's routing formulation is one concrete way to make that choice explicit and testable rather than implicit, and the resulting 64\% reduction in relational knowledge-acquisition calls indicates that unconditional retrieval is unnecessary far more often than standard KGML practice would assume.

\section{Conclusions and Future Work}
\label{sec:conclusion}

This paper introduced Minimum-Sufficient Knowledge and Computation (MSKC), which treats sensing, knowledge, computation, and execution as a jointly routable resource governed by certified feasibility, cost dominance, and break-even guarantees, rather than as a fixed pipeline. We tested its narrowest case, adaptive knowledge acquisition, on real wind ramp-event forecasting using publicly available real data from two farms. The key takeaways are:

\begin{itemize}
\item A state-aware gate cut relational knowledge-acquisition calls by 64\% with no statistically detectable loss in decision quality.
\item The gate beat random gating and matched a simple heuristic.
\item A physics-informed route improved decision quality at essentially no communication cost.
\item A lighter classifier improved severe-class recall, at some cost in macro-F1.
\item Together, these results support MSKC's central premise: uniformly maximal knowledge and computation are not required for every state, and the states where a cheaper route suffices can be identified without a measurable quality loss.
\end{itemize}

Open directions include extending MSKC to the untested sensing, model-fidelity, and execution-placement dimensions; building knowledge routes systematically, dynamically adapting edge knowledge bases rather than hand-crafting each one; and validating portability with more sites. The central open question is whether the feasibility and cost-dominance guarantees established here transfer beyond wind energy to a second resource-constrained cyber-physical domain.
\section*{Acknowledgements}

\footnotesize
Aparna Varde acknowledges NNF Grant NF25OC0110035 from Denmark in the area of Sustainable AI for Environment and Health. This grant supports Purbak Sengupta in his PhD research.

\end{document}